\documentclass[authoryear,preprint,review,12pt]{elsarticle}
\usepackage{cite}
\usepackage{amsmath}
\usepackage{amsfonts}
\usepackage{algorithm}
\usepackage{algorithmic}
\usepackage{threeparttable}
\usepackage{graphicx}
\usepackage{bm}
\usepackage{float}
\usepackage{booktabs}
\usepackage{makecell}
\usepackage{multirow}
\usepackage{pifont}
\usepackage{amssymb}
\usepackage{mathrsfs}
\usepackage{tablefootnote}
\usepackage{soul}
\usepackage{ulem}
\usepackage{subfigure}
\usepackage{comment}
\usepackage{url}
\usepackage[table,xcdraw,dvipsnames]{xcolor}  
\usepackage{color}
\definecolor{citecolor}{RGB}{66,168,235}
\definecolor{linkcolor}{RGB}{255,0,0}
\usepackage{hyperref}
\hypersetup{colorlinks=true,citecolor=citecolor,linkcolor=linkcolor,urlcolor=Thistle}

\usepackage{colortbl}

\makeatother
\journal{ISPRS Journal of Photogrammetry and Remote Sensing}

\biboptions{authoryear}

\begin{document}
	
	\begin{frontmatter}

		\title{Background Debiased SAR Target Recognition via Causal Interventional Regularizer}
		
		\author[address1,address2]{Hongwei Dong\corref{mycorrespondingauthor}}
		\author[address3]{Fangzhou Han\corref{mycorrespondingauthor}}
		\author[address1,address2]{Lingyu Si}
		\author[address1,address2]{Wenwen Qiang}
		\author[address3]{\\Lamei Zhang}
		
		\cortext[mycorrespondingauthor]{Equal Contributions}
		\address[address1]{Science \& Technology on Integrated Information System Laboratory, Institute of Software, Chinese Academy of Sciences, Beijing, China}
		\address[address2]{University of Chinese Academy of Sciences, Beijing, China}
		\address[address3]{Department of Information Engineering, Harbin Institute of Technology, Harbin, China}

		\begin{abstract}
			Recent studies have utilized deep learning (DL) techniques to automatically extract features from synthetic aperture radar (SAR) images, which shows great promise for enhancing the performance of SAR automatic target recognition (ATR). However, our research reveals a previously overlooked issue: SAR images to be recognized include not only the foreground (i.e., the target), but also a certain size of the background area. When a DL-model is trained exclusively on foreground data, its recognition performance is significantly superior to a model trained on original data that includes both foreground and background. This suggests that the presence of background impedes the ability of the DL-model to learn additional semantic information about the target. To address this issue, we construct a structural causal model (SCM) that incorporates the background as a confounder. Based on the constructed SCM, we propose a causal intervention based regularization method to eliminate the negative impact of background on feature semantic learning and achieve background debiased SAR-ATR. The proposed causal interventional regularizer can be integrated into any existing DL-based SAR-ATR models to mitigate the impact of background interference on the feature extraction and recognition accuracy. Experimental results on the Moving and Stationary Target Acquisition and Recognition (MSTAR) dataset indicate that the proposed method can enhance the efficiency of existing DL-based methods in a plug-and-play manner.
		\end{abstract}
		

		
		\begin{keyword}
			Deep learning\sep
			Causal inference\sep
			Synthetic aperture radar\sep
			Automatic target recognition\sep
			Background debias
		\end{keyword}
		
	\end{frontmatter}

	\section{Introduction}
	Synthetic aperture radar (SAR) is an active remote sensor that enables all-weather and day-and-night detection, making it a crucial component of Earth observation systems \citep{moreira2013tutorial}. One of the primary objectives of SAR image data processing is automatic target recognition (ATR) \citep{el2016automatic}. Achieving this goal requires automated and intelligent methods that can accurately recognize targets by extracting discriminative features from SAR images, which remains a challenging task \citep{9374668}. 
	
	Deep learning (DL) overcomes the limitations of traditional manual feature engineering and delivers superior performance in a data-driven manner, making it a powerful solution to address the aforementioned challenge \citep{lecun2015deep}. Researchers have made initial forays into DL-based SAR-ATR methods in recent years \citep{7460942}, but further refinement is necessary before they can be effectively applied to real-world scenarios due to their relatively short development time \citep{zhu2021deep}.
	
	The general structure of SAR-ATR systems consists of three phases: detection, discrimination, and recognition. In related studies, the first two phases, commonly known as the focus-of-attention (FOA), are employed to locate the region of interest (ROI) that may contain targets, without necessitating any human intervention. Both traditional \citep{el2013target} and DL-based approaches \citep{9666902} have demonstrated satisfactory FOA performance.
	
	Using the results of FOA as input, the SAR-ATR method is designed to infer the classes of targets that may be included in the ROI. In this phase, the discriminability of extracted features (i.e., the feature with larger inter-class distance and smaller intra-class distance is more discriminative) is crucial for recognition performance. Various traditional methods have been proposed to improve the feature discriminability, such as hand-crafted feature extractors \citep{amrani2018sar}, attributed scattering centers \citep{li2019sar}, low-rank matrix factorization \citep{zhang2018fusion}. However, DL-based methods offer substantial advantages in terms of performance compared to traditional approaches. \citet{7460942} and \citet{ding2016convolutional} conduct pioneering work that utilize DL-models for SAR image feature extraction and target recognition, setting the foundation for further advancements. Subsequently, a significant portion of research has focused on refining the model architecture to enhance the performance of SAR-ATR, including multi-stream \citep{pei2017sar}, attention mechanism \citep{zhang2020convolutional}, capsule structure \citep{ren2021extended}, vision transformer \citep{9804343}, etc. Given that SAR image processing often encounters the issue of speckle noise, some studies have proposed methods for extracting the speckle-noise-invariant features \citep{8527544,9131844}. Another aspect of the research focuses on adversarial learning, which involves using generative models for data augmentation \citep{sun2019sar} or performing adversarial attacks on existing models \citep{9915465,rs13214358} to enhance the robustness of their feature extraction. In addition, research on DL-based methods for environments with a limited number of training samples \citep{9360316,9515580} and edge devices \citep{9511826} is gradually emerging. 
	\begin{figure}[!ht]
		\begin{center} 
			\subfigure[]{
				\includegraphics[width=0.29\textwidth,height=0.27\textwidth]{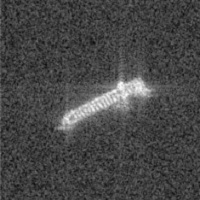}
			}
			\hfil
			\subfigure[]{
				\includegraphics[width=0.29\textwidth,height=0.27\textwidth]{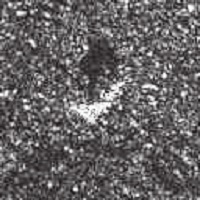}
			}
			\hfil
			\subfigure[]{
				\includegraphics[width=0.29\textwidth,height=0.27\textwidth]{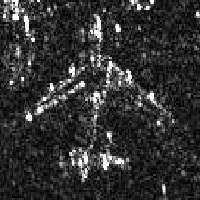}
			}
			\caption{Typical targets in SAR images. (a) Ship target \citep{8067489}. (b) Vehicle target \citep{MSTAR}. (c) Aircraft target \citep{9754573}.}
			\label{fig:targets}
		\end{center}
	\end{figure}
	
	The aforementioned studies have advanced the capability of SAR-ATR from a methodological perspective, but have overlooked the issue of background interference. As shown in Fig. \ref{fig:targets}, the ROI includes not only the foreground (i.e., target), but also a certain size of the background. Therefore, the following question arises:
	\begin{itemize}
		\item[-] \textit{When the foreground and background of two SAR images are different, do the DL-model extracted features belong to the foreground, the background, or a mixture of both?}
	\end{itemize}
	
	A satisfactory answer to this question is that the DL-model extracts semantic information solely from the foreground, as the background is irrelevant for target recognition. In this way, the discriminability of extracted features can be maximized. However, in reality, the situation may differ. Firstly, it should be noted that eliminating the background in ROI can be challenging due to the irregular shape and non-standardized size of non-cooperative targets \citep{8067489,MSTAR,9754573}. Therefore, the input of SAR-ATR typically includes both foreground and background. Then, we conduct a motivating experiment to demonstrate that the features extracted by DL-models are a mixture of foreground and background elements. 
	
\begin{figure}[!hb]
		\begin{center} 
			\subfigure{\label{figmoti_1}}\addtocounter{subfigure}{-1}
			\subfigure[]{\includegraphics[width=0.45\textwidth,height=0.45\textwidth]{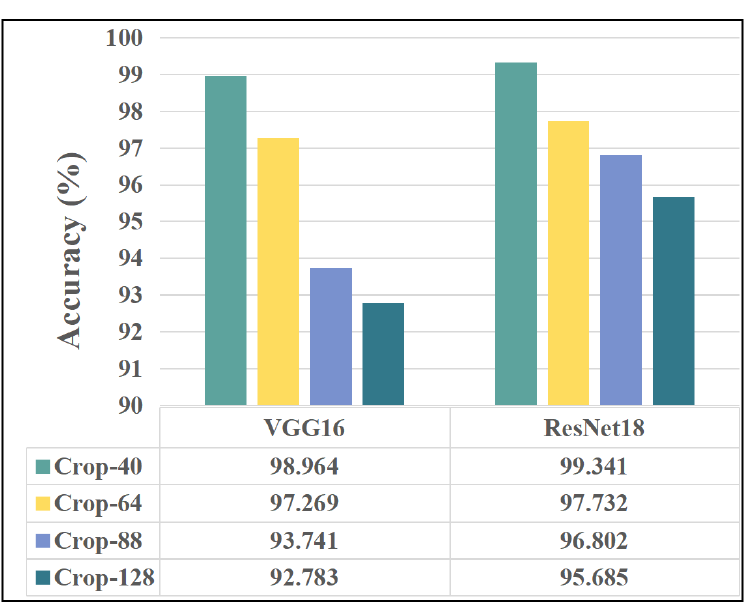}}
			\hspace{0.05cm}
			\subfigure{\label{figmoti_2}}\addtocounter{subfigure}{-1}
			\subfigure[]{\includegraphics[width=0.45\textwidth,height=0.456\textwidth]{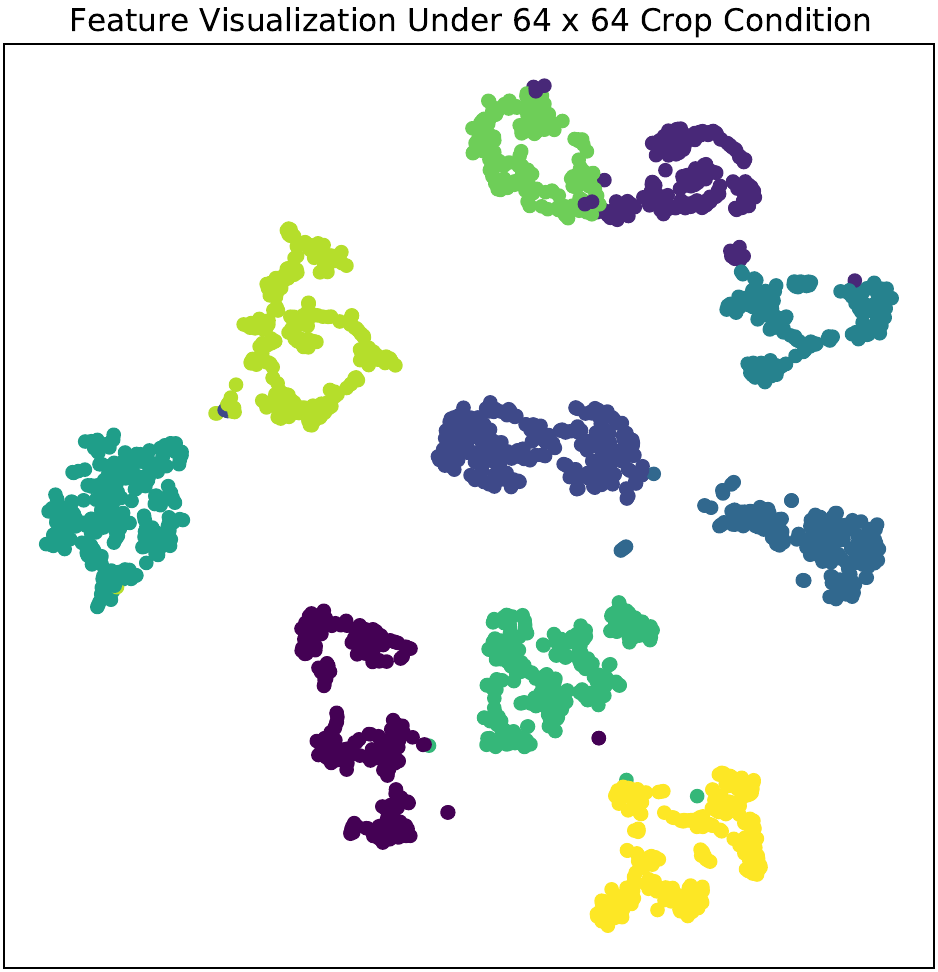}}
			\subfigure{\label{figmoti_3}}\addtocounter{subfigure}{-1}
			\subfigure[]{\includegraphics[width=0.45\textwidth,height=0.456\textwidth]{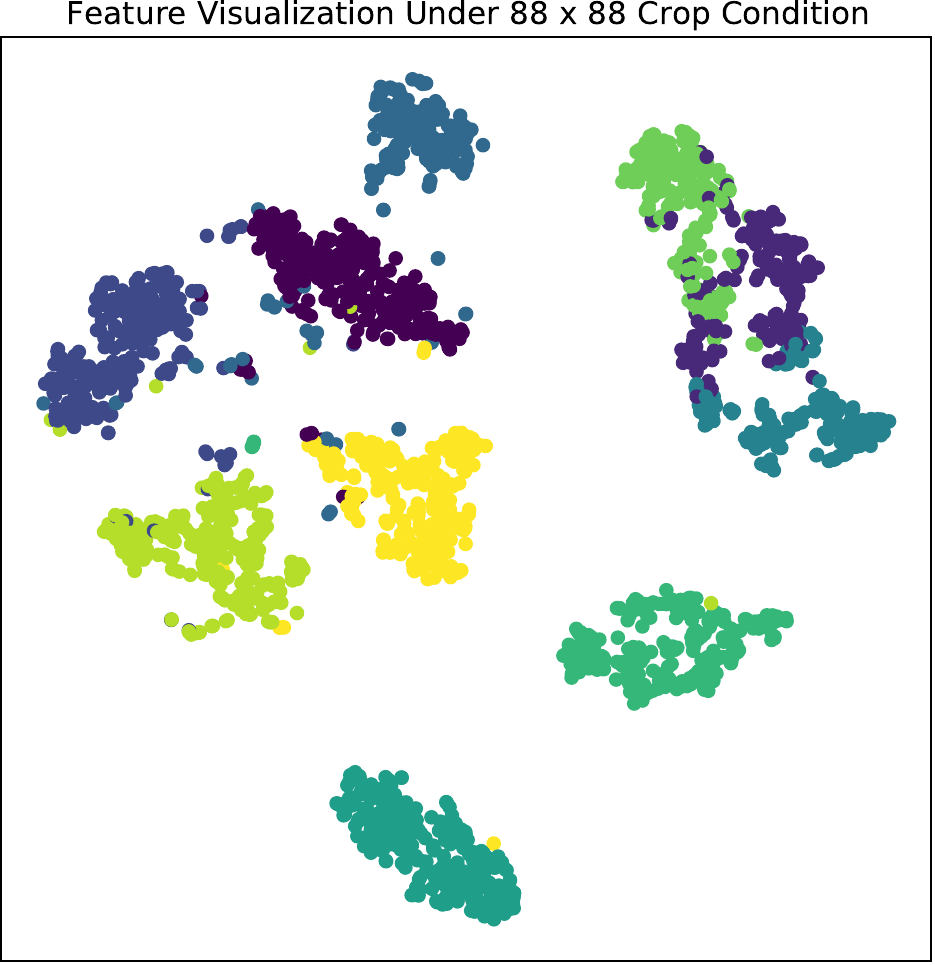}}
			\hspace{0.05cm}
			\subfigure{\label{figmoti_4}}\addtocounter{subfigure}{-1}
			\subfigure[]{\includegraphics[width=0.45\textwidth,height=0.456\textwidth]{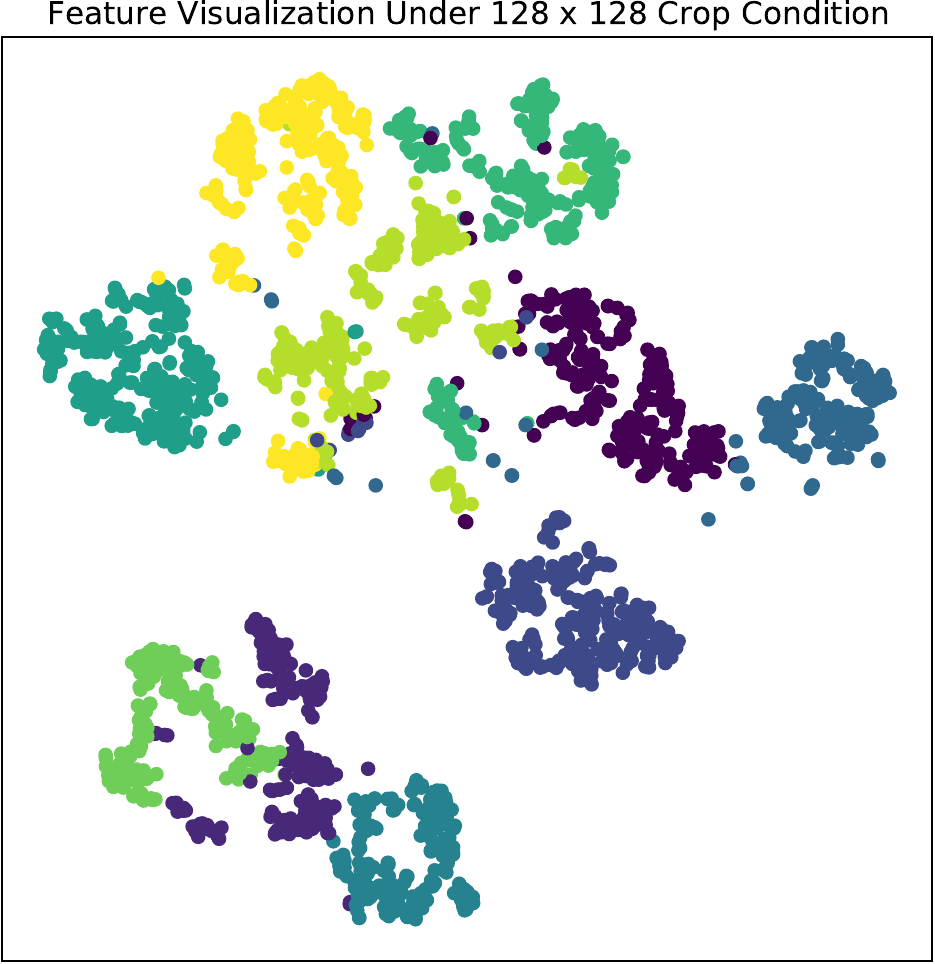}}
			\caption{Results of the motivating experiment for four image crop methods. (a) Recognition accuracy of different center-crops. (b)-(c) Feature visualization of $64\times 64$, $88\times 88$ and $128\times 128$ center-crops using t-SNE \citep{van2008visualizing}.}
			\label{fig:moti}
		\end{center}
	\end{figure}

	Specifically, we employ two DL-based models, i.e., VGG16 \citep{simonyan2014very} and ResNet18 \citep{he2016deep}, and evaluate their performance on the Moving and Stationary Target Acquisition and Recognition (MSTAR) dataset \citep{MSTAR}. This dataset is constructed for scientific research purposes and contains only cooperative vehicle targets that possess similar sizes, roughly measuring 40$\times$40. Based on this observation, we evaluate the performance of two DL-models under four distinct crop conditions: preserving only the central 40$\times$40, 64$\times$64, 88$\times$88, and 128$\times$128 image crops, while masking the remaining areas. The accuracy and feature visualizations of different experimental setups are shown in Fig. \ref{fig:moti}.
	
	As shown in Fig. \ref{fig:moti}(a), the accuracy of SAR-ATR progressively decreases as the image background intensifies. Moreover, Fig. \ref{fig:moti}(b)-(d) demonstrate that the inclusion of background reduces the distance between different classes in the latent space, thereby  diminishing the discriminability of extracted features. This phenomenon is reasonable from a causal perspective since there is no semantic correlation between the foreground and background. Therefore, the elements belonging to the background in the extracted features cannot have a positive impact on target recognition. Both the experimental and theoretical analyses indicate that the background is likely to interfere with the feature extraction process, which should focus on the foreground, leading to biased feature learning objectives and the deterioration of feature quality.
	
	To address this issue, we utilize a structural causal model (SCM) \citep{Causality} to depict the causal relationships among the variables in the SAR-ATR process, including the input, DL-model, foreground, background, and prediction. In the proposed SCM, the background is set as the confounder, which is consistent with the motivating experimental results. By analyzing this SCM, we obtain a theoretical solution for treating background interference. This guides us to represent the theoretical background elimination process as a causal interventional regularizer. As a result, we can achieve background debiased feature learning, support better feature discriminability, and subsequently obtain satisfactory SAR-ATR performance. In summary, the main contributions of this paper are three-fold:
	\begin{itemize}
		\item[(1)] We highlight a critical but overlooked issue that the background exerts an adverse influence on the performance of SAR-ATR, and offer a remedy based on the perspective of causal inference.
		\item[(2)] We construct a SCM based on the process of SAR image feature learning to depict the causal relationships among the input, DL-model, foreground, background, and prediction. This provides a theoretical explanation and solution for background interference.
		\item[(3)] Guided by the constructed SCM, we propose a causal intervention based regularization method. This method can be utilized to effectively mitigate the adverse effects of background on feature learning of the DL-model in a plug-and-play manner, leading to significant enhancement in the performance of SAR-ATR.
	\end{itemize}
	
	The remainder of this paper is organized as follows. Section \ref{sec:2} briefly reviews related works. Section \ref{sec:3} describes the proposed method. Experiment results and analyses are presented in Section \ref{sec:4}. Finally, Section \ref{sec:5} concludes this paper.
	
	\section{Related Works}\label{sec:2}
	This section provides a comprehensive review of the research literature related to this paper, including DL-based SAR-ATR methods, background related methods, and causality theory.
	
	\subsection{General Paradigm of DL-based SAR-ATR}
	Convolutional neural network (CNN) is one of the most significant DL techniques and has achieved state-of-the-art results in various image processing applications, particularly image recognition. The fundamental prototype of CNN, proposed by \citet{Lecun2014Backpropagation}, is characterized by desirable properties such as weight sharing and local connection. \citet{Krizhevsky2012ImageNet} have made significant contributions to the development of current CNN models by using ReLU activation, local response normalization, and GPU training to enhance model capacity and efficiency. CNN is typically regarded as an end-to-end methodology for feature extraction and classification, providing automated feature engineering that is more powerful than traditional hand-crafted features and kernel methods.
	
	The seminal studies of DL-based SAR-ATR method are conducted by \citet{7460942} and \citet{ding2016convolutional}. These studies utilize stacked CNN models composed of convolution, activation, pooling, and fully-connected layers to extract abstract but discriminative deep features from SAR images to achieve promising performance in target recognition. Despite considerable research in this area and some progress being made, the background debiased DL-based SAR-ATR approach has not been explored so far, which is the objective of this paper.
	
	\subsection{Background Debias Related Studies in SAR-ATR}
	Several studies have analyzed the effect of background on SAR-ATR \citep{papson2012classification,cui2005automatic,lombardo2001sar}. Empirical evidence provides by \citet{zhou2019complex} demonstrates the significant impact of background replacement on the performance of SAR-ATR. Moreover, \citet{belloni2020explainability} discuss the role of target, shadow, and clutter in recognition to gain insight into the decision-making process of DL-models.
	
	While there is currently no research exploring background bias from a methodological perspective, the impact of this issue has been observed in the implementation of many SAR-ATR methods. In related studies, the negative effects caused by the background are mitigated by employing two main techniques, i.e., image pre-processing and data augmentation. The image pre-processing method \citep{9454057} involves a multi-stage algorithm that begins by detecting the edges of the target, allowing for the segmentation of the target region from the background. Then, chip recognition is performed to improve performance. On the other hand, the data augmentation method \citep{ding2016convolutional,7460942} is based on sampling a large number of chips from the original image with the same foreground but slightly different backgrounds. These operations prompt the DL-model to learn background-irrelevant features.
	
	However, both of the techniques mentioned above have obvious drawbacks: Firstly, the image pre-processing method not only risks losing target details but also has a high level of complexity. Secondly, the data augmentation method relies on a strong prior, i.e., all targets to be recognized have a fixed and known size. This approach can be challenging when dealing with the recognition problem of non-cooperative targets. In this paper, we address this issue from a causal inference perspective and propose an end-to-end method with a plug-and-play capability.
	
	\subsection{Causality Theory}
	Causality has become as a significant subfield of statistics, with its own conceptual framework, descriptive language, and methodological system. A comprehensive review of the existing literature in \citep{10.1145/3501714.3501755} highlights how causality theory has been instrumental in the development of machine learning by addressing the deficiencies of current approaches. 
	
	SCM \citep{Causality} is a crucial tool in causality theory as it can effectively depict the interdependencies between factors involved in a process using a graph structure. By scrutinizing the SCM, every conceivable alteration that may take place within a process and the pluralistic consequences arising from these alterations may be identified. With the help of SCM, the distinction between correlation and causality can be explored, leading to the current causal inference framework \citep{glymour2016causal}. This framework decomposes the problem into three levels: correlation, intervention, and counterfactual, corresponding to observation, action, and imagination, respectively. Intervention, which involves altering part of the variable generation mechanism while preserving the remainder, is a critical operation for determining and quantifying causality. Scientific research commonly employs intervention, such as through randomized controlled experiments, to establish the presence and direction of causality.
	
	\section{Methodology}\label{sec:3}
	This section presents a detailed introduction to the proposed method for causal intervention based regularization. First, we introduce the construction and analyses of the SAR-ATR oriented SCM. Next, we provide a comprehensive overview of the proposed method, including its model architecture and loss function.
	
	In contrast to many variants in this field that prioritize model refinement, the proposed method focuses on a background deibiased feature learning process. This approach eliminates interference from the background in DL-based SAR-ATR models, guided by causal intervention. This process significantly enhances the discriminability of the extracted features and resulting in superior recognition performance.
	
	\subsection{Causal Analysis}
	To illustrate the impact of background in the process of SAR-ATR, we use the SCM as a basis for our analysis. The SCM is a directed acyclic graph in which nodes represent variables and directed edges indicate causality between them.
	\begin{figure}[!h]
		\centering
		\includegraphics[width=0.85\textwidth]{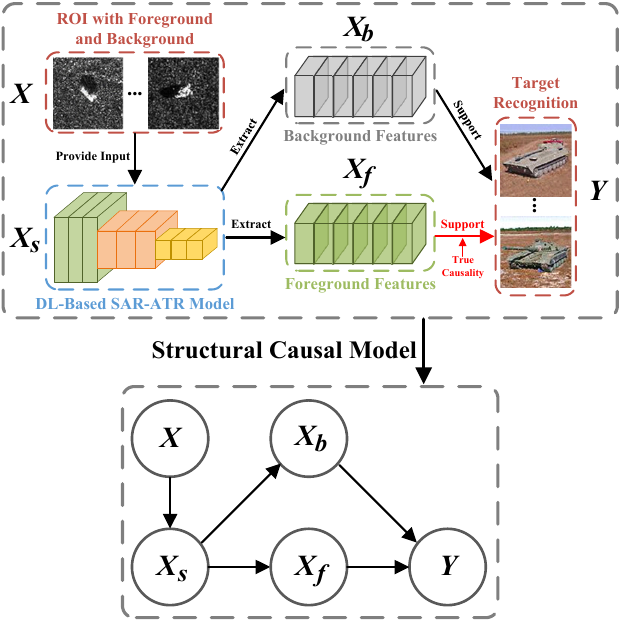}
		\caption{Illustration of the constructed SCM. It depict the causal relationships among the variables in the SAR-ATR process, including the input, DL-model, foreground, background, and prediction. The top part uses examples to demonstrate the conceptual causality. The bottom part is the abstracted SCM for DL-based SAR-ATR.}
		\label{fig:scm}
	\end{figure}
	
	As shown in Fig. \ref{fig:scm}, we refer to the input SAR images as $\bm{X}$, the feature extractor of the DL-based SAR-ATR model as $\bm{X_s}$, and the features extracted by $\bm{X_s}$ as $(\bm{X_f}; \bm{X_b})$, where $\bm{X_f}$ corresponds to foreground features and $\bm{X_b}$ corresponds to background features. The recognition of the label $\bm{Y}$ is supported by both $\bm{X_f}$ and $\bm{X_b}$. In the following, we provide a detailed description of the proposed SCM and the underlying principles behind its construction.
	
	$\bm{X}\rightarrow\bm{X_s}$ indicates that the DL-model $\bm{X_s}$ is used to extract features from the input SAR images $\bm{X}$. $\bm{X_s}\rightarrow\bm{X_b}$ indicates that background features $\bm{X_b}$ is extracted by the model $\bm{X_s}$. $\bm{X_s}\rightarrow\bm{X_f}$ indicates that $\bm{X_f}$ is extracted by $\bm{X_s}$. $\bm{X_b}\rightarrow\bm{Y}$: This link indicates that recognition of input SAR images relies on the information in the background features $\bm{X_b}$, since all  DL-model extracted features are used for the prediction of $\bm{Y}$. $\bm{X_f}\rightarrow\bm{Y}$: This link represents that the label of an input sample is determined by its foreground features, which is the true causality.
	
	As previously analyzed, the ideal DL-model should capture the true causality between $\bm{X_f}$ and $\bm{Y}$. Therefore, we expect the prediction of the input SAR image to be driven primarily by the foreground features, rather than the background features. However, as shown in the constructed SCM, the conventional DL-model fails to capture such causality because the prediction $\bm{Y}$ is not only determined by $\bm{X_f}$ via the link $\bm{X_f}\rightarrow\bm{Y}$, but also by the spurious correlation via the link $\bm{X_f}\leftarrow\bm{X_s}\rightarrow\bm{X_b}\rightarrow\bm{Y}$. It means that the DL-based feature extractor $\bm{X_s}$ generates the background features $\bm{X_b}$ from the background of $\bm{X}$. These features provide part of the semantics that are utilized when recognizing targets. 
	
	The above causalities can be reflected in the motivating experiment corresponding to Fig. \ref{fig:moti}: The performance of DL-based SAR-ATR models trained on input SAR images that include both foreground and background is significantly lower compared to models trained on background-removed data. To capture the true causality while suppressing the spurious correlation, we perform backdoor adjustment by intervening on $\bm{X_f}$ using the $\textit{do}(\cdot)$ operation. This allows us to express the causality between $\bm{X_f}$ and $\bm{Y}$ as $P(\bm{Y}|\textit{do}(\bm{X_f}))$, rather than relying on $P(\bm{Y}|\bm{X_f})$.
	
	It is widely recognized that the channels of the extracted features from any DL-based feature extractor are closely related to the underlying feature semantics \citep{zhou2016learning}. Then we can simply assume that the output of the last network layer of $\bm{X_s}$, i.e., $(\bm{X_f}; \bm{X_b})$, can be mapped into a series of feature semantics, denoted as $(\bm{X_f}; \bm{X_b})\mapsto\{F_i\}_{i=1}^{i=n}$, where $F_i\in\mathbb{R}^{n_c}$ represents a stratification of feature semantics. Considering that the confounder $\bm{X_b}$ is contained in $(\bm{X_f}; \bm{X_b})$ and $F$ is the feature semantics mapped from $(\bm{X_f}; \bm{X_b})$, it follows that a portion of each $F_i$ (belonging to the foreground) is informative for target recognition, while another portion (belonging to the background) occupies the position that should belong to the target information. Therefore, the goal of backdoor adjustment should be to emphasize the foreground-related portions, and to suppress the background-related portions of $F$ that are irrelevant and harmful to target recognition. In this way, the true causality between $\bm{X_f}$ and $\bm{Y}$ can be excavated. Formally, the backdoor adjustment based $\textit{do}(\cdot)$ operation can be expressed as follows:
	\begin{equation}
		P(\bm{Y}|\textit{do}(\bm{X_f})) = \sum_{i=1}^{n}P(\bm{Y}|\bm{X_f}, F_i)P(F_i).
		\label{eq:1}
	\end{equation}
	
	This equation presents a theoretical solution to eliminate background interference. In the following subsection, we will delve into the specifics of implementing the terms described in \eqref{eq:1}, and we will also detail our proposed method guided by this theoretical solution.
	
	\subsection{Causal Interventional Regularizer}
	In this subsection, we embody the theoretical background interference elimination process as a causal interventional regularizer. 
	
	Before introducing the proposed method, we note that in conventional DL-based models, the functional implementation of $P(\bm{Y}|\bm{X_f})$ is provided by a supervised learning process with the cross-entropy loss function, as follows:
	\begin{equation}
			P(\bm{Y}|\bm{X_f}) = 
			-\sum_{k=1}^K\mathbb{I}(\bm{Y}==k)\log\frac{\exp c_k((\bm{X_f}; \bm{X_b}))}{\sum_{j=1}^K\exp c_j((\bm{X_f}; \bm{X_b}))}
		\label{eq:2}
	\end{equation}
	where $k$ represents the $k$th class, $c$ is the mapping: $\mathbb{R}^{n_c}\rightarrow\mathbb{R}^K$ from the last network layer to the softmax classification layer, and $j$ is the $j$th dimension of the prediction probability.
	\begin{figure*}[!ht]
		\centering
		\includegraphics[width=0.975\textwidth]{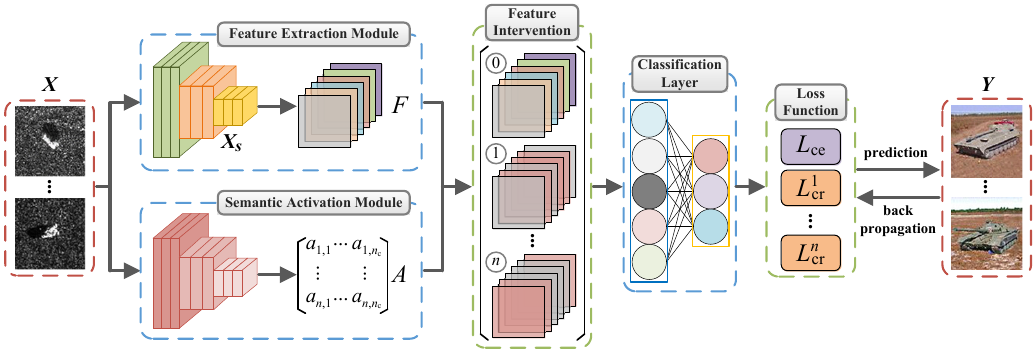}
		\caption{Illustration of the proposed causal intervention based regularization method. It is comprised of two modules, i.e., the feature extraction module and the semantic activation module. The former is a conventional DL-model for extracting the foreground and background features from the input SAR images. The latter can be seen as a causal interventional regularizer that is used to intervene on the obtained feature semantics, thus achieving the background debiased feature learning. In this figure, the instances and labels of training samples, network architectures, and intermediate outputs are marked by red, blue, and green boxes, respectively.}
		\label{fig:ours}
	\end{figure*}
	
	To embody the backdoor adjustment process described in \eqref{eq:1}, the first step is to provide detailed functional implementations for $P(\bm{Y}|\bm{X_f}, F_i)$ and $P(F_i)$. Considering that $F_i$ contains portions belonging to both foreground and background, generating large weights for the foreground-related channels and small weights for the background-related channels of $F_i$ is necessary. By re-weighting $F_i$ based on these weights, we can effectively address the need for background interference elimination. 
	
	Based on the above analyses, when a weight vector $\alpha_i=\left[\alpha_{i,1},\alpha_{i,2},\cdots,\alpha_{i,n_c}\right]^\text{T}$ is given, the  functional implementation for the first term of \eqref{eq:1} can be expressed as:
	\begin{equation}
			P(\bm{Y}|\bm{X_f}, F_i) = 
			-\sum_{k=1}^K\mathbb{I}(\bm{Y}==k)\log\frac{\exp c_k(F_i\odot\alpha_i)}{\sum_{j=1}^K\exp c_j(F_i\odot\alpha_i)}
		\label{eq:3}
	\end{equation}
	where $F_i\odot\alpha_i=\left[F_{i,1}\cdot\alpha_{i,1},F_{i,2}\cdot\alpha_{i,2},\cdots,F_{i,n_c}\cdot\alpha_{i,n_c}\right]$. Therefore, given a weight matrix $A=\left[\alpha_1,\alpha_2,\cdots,\alpha_n\right]$, and let $P(F_i)=1/n$, the overall implementation of the backdoor adjustment based $\textit{do}(\cdot)$ operation can be expressed as:
	\begin{equation}
			P(\bm{Y}|\textit{do}(\bm{X_f})) = 
			-\sum_{i=1}^n(\sum_{k=1}^K\mathbb{I}(\bm{Y}==k)\log\frac{\exp c_k(F_i\odot\alpha_i)\times\frac{1}{n}}{\sum_{j=1}^K\exp c_j(F_i\odot\alpha_i)}).
		\label{eq:4}
	\end{equation}
	
	The aforementioned approach shifts our focus from providing functional implementations for \eqref{eq:1} to the task of generating the weight matrix $A$ required in \eqref{eq:4}. To accomplish this, we use a learnable network to generate the required weight matrix. This idea inspires us to propose a novel regularization method with plug-and-play capability based on causal intervention. The DL-model that incorporates the proposed causal interventional regularizer is comprised of two modules, i.e., the feature extraction module and the semantic activation module. The unified framework for this approach is depicted in Fig. \ref{fig:ours}. In the following, we give detailed descriptions of the two modules and the functional learning objective.
	
	\subsubsection{Feature Extraction Module}
	When integrating the proposed method with DL-based SAR-ATR models, a conventional DL-based feature extractor is utilized as the backbone to extract \textit{task-specific features} from the SAR images containing both foreground and background. Therefore, the backbone of this module corresponds to $\bm{X_s}$ in Fig. \ref{fig:scm}, while its output corresponds to the feature semantics of $(\bm{X_f}; \bm{X_b})$. 
	
	We denote the training set as $D_\text{train}=\{X^i,Y^i\}_{i=1}^{i=N}$, where $X^i\in\mathbb{R}^{w\times h\times c}$ represents the instance with a size of $w\times h\times c$, and $Y^i\in\mathbb{R}$ represents the label in this set. The forward propagation of the feature extraction module can be expressed as $f_\text{sem}(X^i)$, where $f_\text{sem}(\cdot)$ is the mapping of the backbone of this module, parameterized by $W_\text{sem}$. It is used to map the input to the semantic space, i.e., $f_\text{sem}: \mathbb{R}^{w\times h\times c}\mapsto\mathbb{R}^{n\times n_c}$.
	
	The proposed method has a plug-and-play capability, as we do not impose any restrictions on the feature extractor. This is evident from the definition of $f_\text{sem}(\cdot)$, allowing any DL-model to be used to build the feature extraction module.
	
	\subsubsection{Semantic Activation Module}
	After obtaining the preliminary feature semantics $F$ using the feature extraction module, another backbone is utilized by this module to extract the \textit{semantic-activation-specific features} and generate the weight matrix in \eqref{eq:4}. Subsequently, the process of re-weighting from the preliminary to the background debiased feature semantics, i.e., $F_i\odot\alpha_i$, can be performed by intervening the feature semantics.
	
	The forward propagation of the semantic activation module can be expressed as $f_\text{sem}(X^i)\odot f_\text{sam}(X^i)$, where $f_\text{sam}(\cdot)$ is the mapping of the backbone of this module, parameterized by $W_\text{sam}$. It is used to map the input to the semantic activation space, i.e., $f_\text{sam}: \mathbb{R}^{w\times h\times c}\mapsto\mathbb{R}^{n\times n_c}$, which activates the feature semantics belonging to the foreground while suppressing those belonging to the background, thereby excavating the true causality.
	
	\subsubsection{Overall Learning Objective}
	The classification layer, parameterized by $W_\text{cls}$, is used to map the preliminary and background debiased feature semantics to prediction probabilities. Then the forward prediction and loss back-propagation of our proposed framework can be supported. Following the definitions in \eqref{eq:2} to \eqref{eq:4}, for the training set $D_\text{train}$, the learning objective for the feature extraction module can be expressed as the following cross-entropy loss:
	\begin{equation}
		\begin{split}
			&L_\text{ce}(X,Y,W_\text{sem},W_\text{cls}) = \\
			&-\frac{1}{N}\sum_{i=1}^N\sum_{k=1}^K\mathbb{I}(Y^i==k)\log\frac{\exp c_k(f_\text{sem}(X^i))}{\sum_{j=1}^K\exp c_j(f_\text{sem}(X^i))}.
		\end{split}
		\label{eq:lce}
	\end{equation}
	
	Besides, the $t$th learning objective for the semantic activation module can be expressed as the following causal interventional regularizer loss:
	\begin{equation}
		\begin{split}
			&L_\text{cr}^t(X,Y,W_\text{sem},W_\text{sam},W_\text{cls}) = \\&-\frac{1}{N}\sum_{i=1}^N\sum_{k=1}^K\mathbb{I}(Y^i==k)
			\log\frac{\exp c_k(f_\text{sem}^t(X^i)\odot f_\text{sam}^t(X^i))\times \frac{1}{n}}{\sum_{j=1}^K\exp c_j(f_\text{sem}^t(X^i)\odot f_\text{sam}^t(X^i))},
		\end{split}
		\label{eq:lcr}
	\end{equation}
	and the overall learning objective for this module is the sum of $n$ times calculation for \eqref{eq:lcr}. Based on \eqref{eq:lce} and \eqref{eq:lcr}, we can obtain the overall learning objective of the unified framework shown in Fig. \ref{fig:ours}:
	\begin{equation}
		L_\text{total} = L_\text{ce} + \lambda\sum_{t=1}^n L_\text{cr}^t 
		\label{eq:overall}
	\end{equation}
	where $\lambda$ is the hyperparameter to control the trade-off. The two terms of \eqref{eq:overall} correspond to the $P(\bm{Y}|\bm{X_f})$ and $P(\bm{Y}|\textit{do}(\bm{X_f}))$ in the theoretical causal analysis, which reflects the integration of the conventional DL-model with our proposed causal interventional regularizer.
	
	It is important to note that in a conventional DL-based SAR-ATR model, the loss function only includes the $L_\text{ce}$ term. Therefore, the proposed method can be seen as a way to regularize the conventional model using the causal inference induced $L_\text{cr}$ term. That is why we refer to the proposed method as \textit{causal interventional regularizer} in this paper.
	
	The training process of the proposed method is outlined as Algorithm \ref{alg:1}.
	
	\begin{algorithm}
		\caption{Training the Background Debiased SAR-ATR Model}
		\begin{algorithmic}[1]
			\REQUIRE ~~\\
			Model parameter initialization $W_{\text{sem}}$, $W_{\text{sam}}$, and $W_{\text{cls}}$; \\
			Epoch $I$;\\
			Mini-batch size $B$;\\
			Learning rate $\eta$;\\
			Hyperparameter $\lambda$;\\
			Training and validation sets $D_\text{train}$ and $D_\text{val}.$
			\FOR{epoch in $I$:}
			\FOR{batch in $B$:}
			\STATE Forward propagation by $f_{\text{sem}}$, $f_{\text{sam}}$, and $c$ with trainable parameters of $W_{\text{sem}}$, $W_{\text{sam}}$, and $W_{\text{cls}}$
			\STATE \# update the classification layer
			\STATE $W_{\text{cls}}\leftarrow W_{\text{cls}} - \eta\nabla_{W_{\text{cls}}}L_\text{total}$
			\STATE \# update the semantic activation module
			\STATE $W_{\text{sam}}\leftarrow W_{\text{sam}} - \eta\nabla_{W_{\text{sam}}}L_\text{total}$
			\STATE \# update the feature extraction module
			\STATE $W_{\text{sem}}\leftarrow W_{\text{sem}} - \eta\nabla_{W_{\text{sem}}}L_\text{total}$
			\ENDFOR
			\ENDFOR
			\ENSURE ~~\\
			The DL-model with the highest validation accuracy.
		\end{algorithmic}
		\label{alg:1}
	\end{algorithm}
	
	\section{Experiments}\label{sec:4}
	In this section, we first introduce the dataset used in the experiments. Then we proceed to demonstrate the performance of both the proposed method and the compared methods. Finally, we analyze the effect of the hyperparameter in the proposed method on its target recognition performance.
	
	\subsection{MSTAR Dataset}
	The experimental data utilized in this paper is obtained through the employment of the Sandia National Laboratory SAR sensor platform. This venture received funding from the Defense Advanced Research Projects Agency and the Air Force Research Laboratory, as an integral component of the MSTAR program \citep{MSTAR}. The dataset released to the public comprise ten distinct ground target classes, namely armored personnel carrier types BMP-2, BRDM-2, BTR-60, and BTR-70; tanks T-62 and T-72; rocket launcher 2S1; air defense unit ZSU-234; truck ZIL-131; and bulldozer D7. The data is collected using an X-band SAR sensor, in a spotlight mode with a resolution of 1-ft and full aspect coverage (spanning $0^{\circ}$ to $360^{\circ}$).
	\begin{figure*}[!ht]
		\centering
		\subfigure[2S1]{
			\begin{minipage}[b]{0.13\textwidth}
				\includegraphics[width=1\linewidth, height=0.875\linewidth]{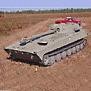}\vspace{4pt}
				\includegraphics[width=1\linewidth, height=0.875\linewidth]{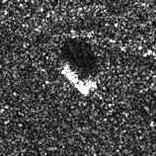}
		\end{minipage}}
		\hspace{0.75em}
		\subfigure[BMP-2]{
			\begin{minipage}[b]{0.13\textwidth}
				\includegraphics[width=1\linewidth, height=0.875\linewidth]{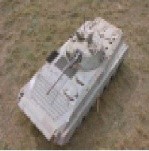}\vspace{4pt}
				\includegraphics[width=1\linewidth, height=0.875\linewidth]{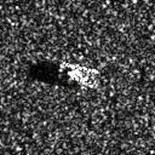}
		\end{minipage}}
		\hspace{0.75em}
		\subfigure[BRDM-2]{
			\begin{minipage}[b]{0.13\textwidth}
				\includegraphics[width=1\linewidth, height=0.875\linewidth]{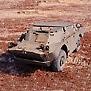}\vspace{4pt}
				\includegraphics[width=1\linewidth, height=0.875\linewidth]{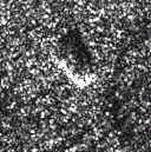}
		\end{minipage}}
		\hspace{0.75em}
		\subfigure[BTR-60]{
			\begin{minipage}[b]{0.13\textwidth}
				\includegraphics[width=1\linewidth, height=0.875\linewidth]{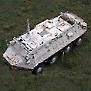}\vspace{4pt}
				\includegraphics[width=1\linewidth, height=0.875\linewidth]{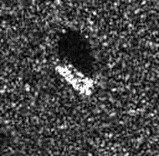}
		\end{minipage}}
		\hspace{0.75em}
		\subfigure[BTR-70]{
			\begin{minipage}[b]{0.13\textwidth}
				\includegraphics[width=1\linewidth, height=0.875\linewidth]{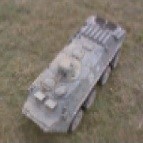}\vspace{4pt}
				\includegraphics[width=1\linewidth, height=0.875\linewidth]{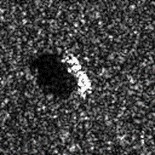}
		\end{minipage}}
		
		\subfigure[D7]{
			\begin{minipage}[b]{0.13\textwidth}
				\includegraphics[width=1\linewidth, height=0.875\linewidth]{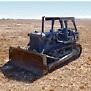}\vspace{4pt}
				\includegraphics[width=1\linewidth, height=0.875\linewidth]{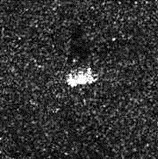}
		\end{minipage}}
		\hspace{0.75em}
		\subfigure[T-62]{
			\begin{minipage}[b]{0.13\textwidth}
				\includegraphics[width=1\linewidth, height=0.875\linewidth]{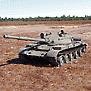}\vspace{4pt}
				\includegraphics[width=1\linewidth, height=0.875\linewidth]{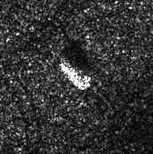}
		\end{minipage}}
		\hspace{0.75em}
		\subfigure[T-72]{
			\begin{minipage}[b]{0.13\textwidth}
				\includegraphics[width=1\linewidth, height=0.875\linewidth]{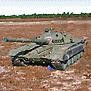}\vspace{4pt}
				\includegraphics[width=1\linewidth, height=0.875\linewidth]{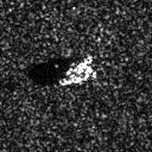}
		\end{minipage}}
		\hspace{0.75em}
		\subfigure[ZIL-131]{
			\begin{minipage}[b]{0.13\textwidth}
				\includegraphics[width=1\linewidth, height=0.875\linewidth]{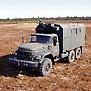}\vspace{4pt}
				\includegraphics[width=1\linewidth, height=0.875\linewidth]{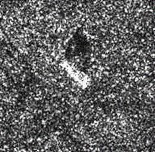}
		\end{minipage}}
		\hspace{0.75em}
		\subfigure[ZSU-234]{
			\begin{minipage}[b]{0.13\textwidth}
				\includegraphics[width=1\linewidth, height=0.875\linewidth]{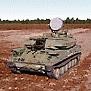}\vspace{4pt}
				\includegraphics[width=1\linewidth, height=0.875\linewidth]{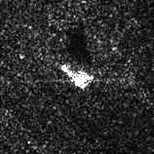}
		\end{minipage}}
		\caption{Ten types of military targets in the MSTAR dataset in the optical and SAR formats. (a) 2S1. (b) BMP-2. (c) BRDM-2. (d) BTR-60. (e) BTR-70. (f) D7. (g) T-62. (h) T-72. (i) ZIL-131. (j) ZUS-234.}
		\label{fig:mstar}
	\end{figure*}
	
	The MSTAR benchmark dataset is widely employed for evaluating the performance of SAR-ATR methods. Fig. \ref{fig:mstar} illustrates the SAR images of ten target classes contained in the MSTAR dataset, captured at similar aspect angles. In addition, it displays their corresponding optical images. In our experiments, involved SAR-ATR methods are thoroughly evaluated under standard operating conditions (SOC) as well as extended operating conditions (EOC) to ensure their appropriate evaluation. SOC primarily includes utilizing serial numbers and target configurations in the testing set that align with those in the training set, however, with varying aspects and depression angles. On the other hand, EOC involves substantial differences between the training and testing sets, highlighting notable variations in depression angle, target articulation, and version variants.
	
	\subsection{Experimental Settings}
	The effectiveness of the proposed method is demonstrated through comparisons with notable alternatives, including VGG16 \citep{simonyan2014very}, ResNet18 \citep{he2016deep}, and A-ConvNet \citep{7460942}. The former two are the predominant DL-models utilized in image recognition, whereas A-ConvNet holds extensive application in SAR-ATR domain. 
	
	In the experiments, we use a VGG16 model as the backbone of the semantic activation module. The patch size of input images is limited to 128$\times$128. The size of obtained feature map from backbone is 4$\times$4$\times$512. When integrating the proposed method with the comparison methods, they will be used as the feature extractor in the feature extraction module in the proposed framework, and a fully-connected layer is attached behind them to generate the feature semantics.
	
	Sophisticated data pre-processing and augmentation techniques are frequently employed in related studies for better recognition performance. Nevertheless, to illustrate the inherent target recognition proficiency of the DL-models, we refrain from carrying out any pre-processing activities, such as de-speckle, on the SAR images to be recognized in the experimental setup. Furthermore, to accentuate the background debias capability of the proposed method, we do not perform any data augmentation for VGG16 and ResNet18 models, but retain the original data augmentation operation of A-ConvNet, which involves randomly sampling a substantial quantity of $88\times 88$ patches from the original $128\times 128$ SAR images. This data augmentation method aims to mitigate the negative impact of background on the feature learning of targets. The aforementioned settings facilitate a straightforward comparison between the effectiveness of the proposed and the data augmentation methods for the issue of background debias.
	
	The mini-batch stochastic gradient descent based method is used for model optimization, with a batch size of $64$, and a training epoch of $100$. In order to accelerate the optimization process and obtain more precise approximate solutions, the Adam optimizer is employed with the learning rate of $0.01$. Upon completion of the training process, the model with the highest validation accuracy is preserved.
	
	The experiments of this study employ a NVIDIA RTX A6000 GPU, and the program is implemented using the PyTorch DL framework.
	
	\subsection{Experiments under SOC}
	In the experimental setup of SOC, which is listed in Table \ref{tab:1}, the involved methods are evaluated on the ten-class recognition problem. 
	\begin{table}[!ht]
		\centering
		\caption{Number of Training and Testing samples for the SOC Experimental Setup}
		\resizebox{0.65\textwidth}{!}
		{
			\renewcommand\arraystretch{1.5}
			\begin{tabular}{|c|c|c|}
				\hline
				Class&Training Set $(17^{\circ})$&Testing Set $(15^{\circ})$\\ \hline
				2S1&299&274\\ \hline
				BMP-2&233&196\\ \hline
				BRDM-2&298&274\\ \hline
				BTR-60&256&195\\ \hline
				BTR-70&233&196\\ \hline
				D7&299&274\\ \hline
				T-62&299&273\\ \hline
				T-72&232&196\\ \hline
				ZIL-131&299&274\\ \hline
				ZIL-234&299&274\\ \hline
			\end{tabular}
		}
		\label{tab:1}
	\end{table}
	\begin{table*}[!ht]
		\centering
		\caption{The comparison of Recognition Accuracy (\%) under SOC}
		\resizebox{0.95\textwidth}{!}
		{
			\renewcommand\arraystretch{1.5}
			\begin{tabular}{|c|c|c|c|c|}
				\hline
				&CNN&VGG16&ResNet18&A-ConvNet \\
				\hline
				\multirow{3}*{Accuracy(\%)} 
				&91.483 & 92.783&95.685&98.376 \\ \cline{2-5}
				&CNN+Ours &VGG16+Ours&ResNet18+Ours&A-ConvNet+Ours \\ \cline{2-5}
				&95.637 & 98.557&98.215&99.228 \\
				\hline
			\end{tabular}
		}
		\label{tab:2}
	\end{table*}
	\begin{figure*}[!ht]
		\centering
		\subfigure[]{\includegraphics[width=0.45\textwidth]{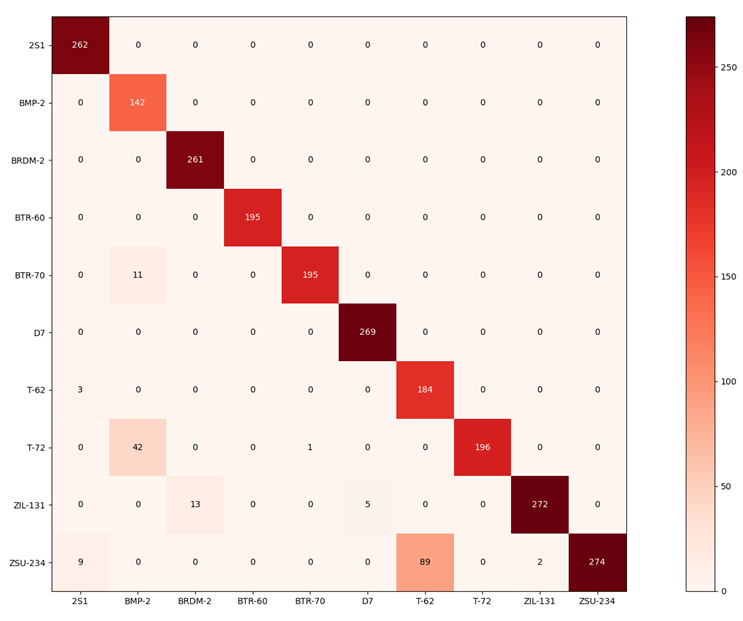}\label{fig:confusion1a}}
		\hspace{2em}
		\subfigure[]{\includegraphics[width=0.45\textwidth]{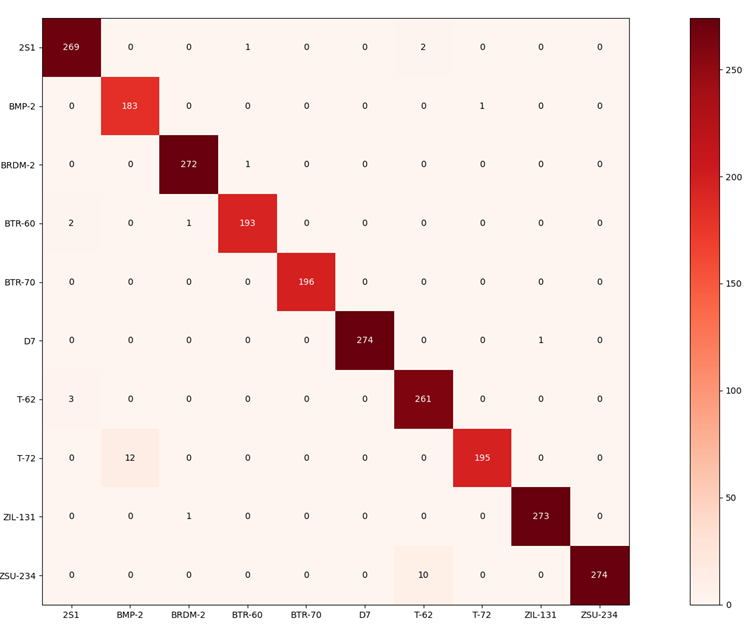}\label{fig:confusion1b}}    
		\caption{The comparison of confusion matrices under SOC. (a) Results of VGG16. (b) Results of background debiased VGG16.}
		\label{fig:confusion}
	\end{figure*}
	
	It must be noted that the target serial number is identical in both the training and testing sets, although they vary in their azimuth and depression angles. The training images are acquired at a depression angle of $17^{\circ}$, whereas the testing images are captured with a depression angle of $15^{\circ}$.
	
	Table \ref{tab:2} shows the comparison of the testing accuracy of CNN, VGG16, ResNet18, A-Convnet before and after adding the proposed causal interventional regularizer. In addition, the comparison of confusion matrices before and after background debias for the VGG16 model is shown in Fig. \ref{fig:confusion}.
	
	As can be seen from the experimental results, for the three DL-models without data augmentation, they also show significant improvements in the testing accuracy after adding the proposed causal interventional regularizer. Among them, background debiased VGG16 model even outperforms the A-ConvNet model, which expands the original training set 1681 times by a random clipping based data augmentation method. Specifically, after the causal-driven regularizer is inserted, the involved four DL-models have 4.154\%, 5.774\%, 2.530\%, and 0.852\% improvement in the recognition accuracy, respectively. The greatly improved SAR-ATR performance shows that even if there are many background areas in the ROI that interfere with target semantic learning, the conventional DL-based models can still effectively focus on the extraction of foreground information after adding the proposed causal interventional regularizer. This demonstrates the satisfactory background debiased capability of the proposed method. In addition, since the proposal is a plug-and-play method, there is no need to make any modifications to the original model, which greatly improves the generalizability of various DL-based recognition models in the field of SAR-ATR.
	
	\subsection{Experiments under EOC}
	It is widely acknowledged that SAR images are highly susceptible to variations in depression angles. Therefore, the robustness of SAR-ATR methods to the changes in depression angle is important. To this end, we assess the proposed method in terms of large depression angle changes (indicated by EOC-1).
	\begin{table}[!ht]
		\centering
		\caption{Details of EOC-1 Experimental Setup}
		\resizebox{0.95\textwidth}{!}
		{
			\renewcommand\arraystretch{1.5}
			\begin{tabular}{|c|c|c|c|c|c|}
				\hline
				\multicolumn{2}{|c|}{Target Type}&\multicolumn{2}{c|}{Training Set}&\multicolumn{2}{c|}{Testing Set}\\ \hline
				Class&Serial No.&Depression&No. Samples&Depression&No. Samples\\ \hline
				2S1&B01&$17^{\circ}$&299&$30^{\circ}$&288\\ \hline
				BRDM-2&E71&$17^{\circ}$&298&$30^{\circ}$&287\\ \hline
				T-72&A64&$17^{\circ}$&299&$30^{\circ}$&288\\ \hline
				ZSU-234&D08&$17^{\circ}$&299&$30^{\circ}$&288\\ \hline
			\end{tabular}
		}
		\label{tab:3}
	\end{table}
	
	Table \ref{tab:3} enumerates that among the targets in the MSTAR dataset, only four classes (2S1, BRDM-2, T-72, and ZSU-234) possess targets obtained at a depression angle of $30^{\circ}$. Consequently, we evaluate the involved DL-models on these samples, wherein the corresponding training set is the data of these four classes of targets in SOC, i.e., the same targets obtained at a depression angle of $17^{\circ}$. The large disparity in depression angle may result in a dissimilar depiction of same targets in identical postures, thereby augmenting the difficulty of recognition. 
	
\begin{figure}[!ht]
		\centering
		\subfigure[]{\includegraphics[width=0.47\textwidth]{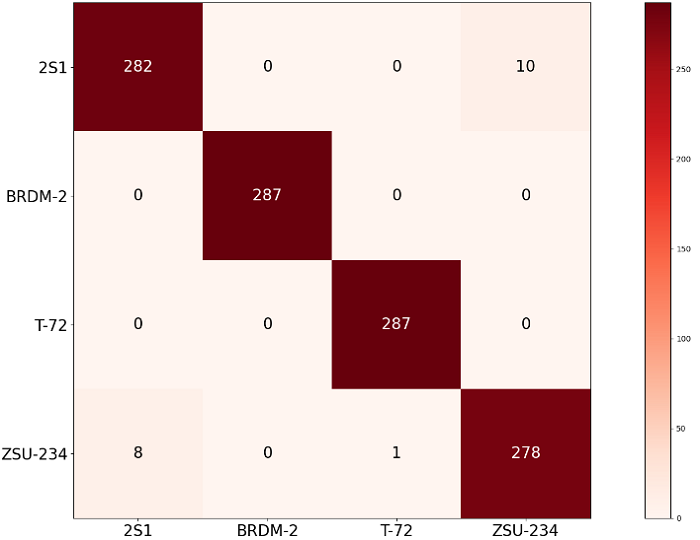}\label{fig:confusion2a}}
		\hspace{0.05em}
		\subfigure[]{\includegraphics[width=0.47\textwidth]{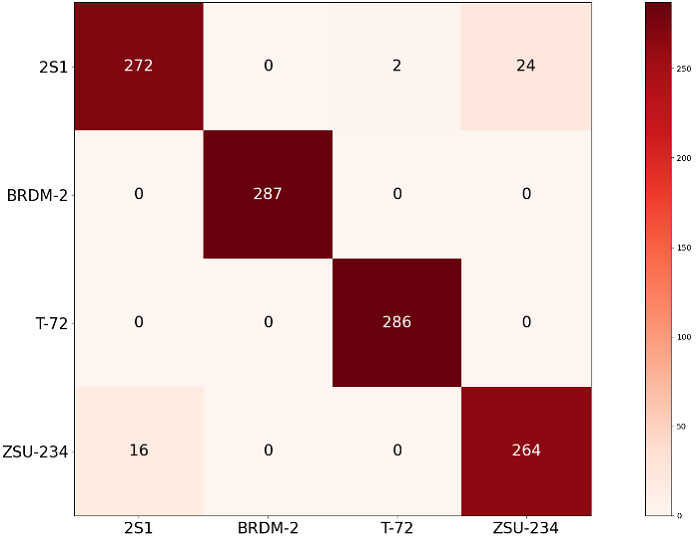}\label{fig:confusion2b}}    
		\caption{The comparison of confusion matrices under EOC-1. (a) Results of VGG16. (b) Results of background debiased VGG16.}
		\label{fig:confusion2}
	\end{figure}

	The EOC-1 dataset is classified using two models: the conventional VGG16 model and the VGG16 model incorporating the proposed causal interventional regularizer. The confusion matrix for large depression angle variations is presented in Fig. \ref{fig:confusion2}. The experimental results indicate that the VGG16 model achieves the accuracy of 96.351\% while our proposed model has a higher accuracy of 98.436\%. This suggests that the proposed method is capable of improving the robustness of DL-based SAR-ATR models even under challenging conditions such as large depression angles.
	
	In another EOC testing scenario, i.e., EOC-2, the assessment of SAR-ATR methods revolves around variations in target configuration. The objective of this validation is to evaluate the performance of the model to recognize different variants of the same target. Specifically, the training set encompasses BMP-2, BRDM-2, BTR-70 and T-72 at depression angles of $15^{\circ}$ and $17^{\circ}$, whereas the testing set comprises of diversified versions of T-72. The variance in target representation between the two sets makes it arduous for the testing samples to be recognized as T-72 targets, thus intensifying the task challenge. Details of EOC-2 experimental setup are shown in Table \ref{tab:4}.
	\begin{table}[!ht]
		\centering
		\caption{Details of EOC-2 Experimental Setup}
		\resizebox{0.75\textwidth}{!}
		{
			\renewcommand\arraystretch{1.5}
			\begin{tabular}{|c|c|c|c|c|}
				\hline
				Partition&Class&Serial No.&Depression&No. Samples\\ \hline
				\multirow{4}*{Training Set}
				&BMP-2&C21&$17^{\circ}$&233\\ \cline{2-5}
				&BRDM-2&E71&$17^{\circ}$&298\\ \cline{2-5}
				&BTR-70&C71&$17^{\circ}$&233\\ \cline{2-5}
				&T-72&132&$17^{\circ}$&232\\ \hline
				\multirow{5}*{Testing Set}
				&\multirow{5}*{T-72}&S7&$15^{\circ}$/$17^{\circ}$&419\\ \cline{3-5}
				& &A32&$15^{\circ}$/$17^{\circ}$&572\\ \cline{3-5}
				& &A62&$15^{\circ}$/$17^{\circ}$&573\\ \cline{3-5}
				& &A63&$15^{\circ}$/$17^{\circ}$&573\\ \cline{3-5}
				& &A64&$15^{\circ}$/$17^{\circ}$&573\\ \hline
			\end{tabular}
		}
		\label{tab:4}
	\end{table}
	
	Experimental results under EOC-2 can be seen from Table \ref{tab:5}. Similar to the previous experiments, we choose VGG16 as a representative of the conventional DL-models to test its change in recognition accuracy before and after the addition of the proposed causal interventional regularizer.
	\begin{table}[!ht]
		\centering
		\caption{The comparison of Recognition Accuracy (\%) under EOC-2}
		\resizebox{0.75\textwidth}{!}
		{
			\renewcommand\arraystretch{1.5}
			\begin{tabular}{|c|c|c|c|}
				\hline
				Serial No.&No. Samples&VGG16&VGG16+Ours\\ \hline
				A32&572&88.462 &95.979 \\ \hline
				A62&573&95.637 &97.906 \\ \hline
				A63&573&94.939 &96.684 \\ \hline
				A64&573&92.845 &95.637 \\ \hline
				S7&419&79.714 &80.191 \\ \hline
				Total&2710&90.923&94.022\\ \hline
			\end{tabular}
		}
		\label{tab:5}
	\end{table}
	
	It is relatively obvious from the experimental results that the incorporation of the proposed method improves the accuracy of the conventional DL-model for the recognition of all variant classes of T-72. In particular, the recognition accuracy of the serial number A32 is improved by 7.517\%, which is a considerable improvement. The above results effectively verify the effectiveness of the proposed method for background debias when recognizing variants of the same target.
	
	\subsection{Hyperparameter Analysis}
\begin{figure}[!ht]
		\centering
		\includegraphics[width=0.85\textwidth]{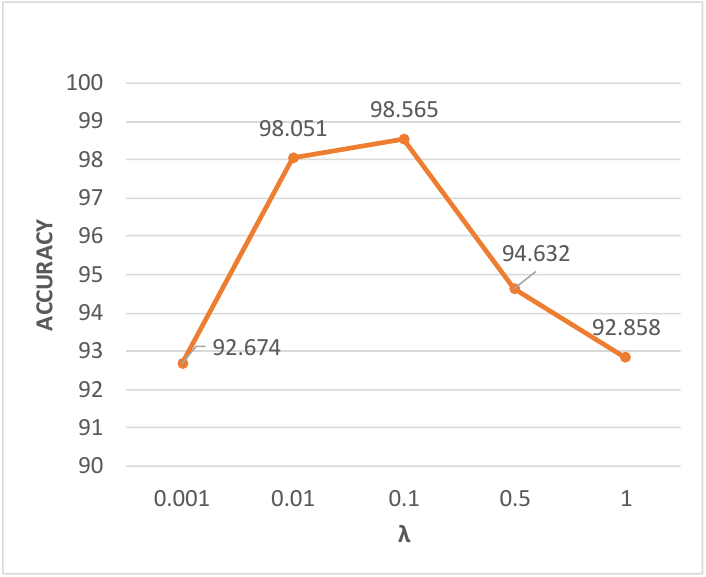}
		\caption{The effect of hyperparameter $\lambda$ on the recognition accuracy.}
		\label{fig:hyper}
	\end{figure}

	To comprehensively investigate the impact of hyperparameter on the SAR-ATR accuracy of the proposed method, we employ the VGG16 model with the proposed regularization method and conduct experiments under SOC. Our primary focus is on assessing the significance of the hyperparameter $\lambda$ in \eqref{eq:overall} on recognition accuracy while moderating the proportion of the causal interventional regularizer induced loss $L_\text{cr}$ in the overall loss $L_\text{total}$, which denotes the intervention degree of the proposed method in the feature extraction of conventional DL-model. We achieve this goal by setting the value of $\lambda$ to $\{0.001, 0.01, 0.1, 0.5, 1.0\}$ and conduct comparative experiments. The results are presented in Fig. \ref{fig:hyper}.
	
	As demonstrated in Fig. \ref{fig:hyper}, the selection of hyperparameter $\lambda$ is critical in promoting a desirable background debias effect. Notably, the highest accuracy is attainable when $\lambda$ is set to $0.1$; deviating from this value, whether by setting it too high or low, can result in a significant decrease in recognition accuracy.
	
	\section{Conclusion}\label{sec:5}
	In this paper, we present the first attempt to investigate the potential of background debias on DL-based SAR-ATR models from a methodological perspective. Specifically, we propose a novel causal intervention based regularization method to eliminate the interference of background on the extraction of target semantic information. The proposed method is developed based on the analysis of the SAR-ATR oriented SCM, which depicts the causal relationships among the input, DL-model, foreground, background, and prediction.
	
	We model the background in the SAR image to be recognized as a confounder for the feature extraction of DL-models, and eliminate it through backdoor adjustment based causal intervention. This theoretical solution for eliminating background interference is then transformed into a regularization term for conventional DL-models, resulting in background debiased feature learning. This method extracts more target semantics and improves the discriminability of the extracted features, while also being easy to implement, low complexity, widely adaptable, and having plug-and-play capability.
	
	Experiments on the MSTAR dataset demonstrate that our proposed method offers advantages over conventional approaches, as evidenced by improved recognition performance when combined with any conventional DL-model. Moving forward, we plan to explore the combination of causal inference and diverse tasks such as SAR target detection and change detection. We will also explore other learning pipelines, including self-supervised learning, few-shot learning, and model compression.

	\section*{CRediT authorship contribution statement}
	
	H.D. and F.H. were responsible for the conceptualization, design and development of the methods, writing programmes, running experiments, analyzing the results, and writing the paper. L.S., W.Q. and L.Z. were responsible for investigation, funding requests and supervision. All authors reviewed the manuscript.
	
	\section*{Conflict of interest}
	
	The authors declare that they have no conflict of interest.

	\section*{Acknowledgments}
	
	This work was supported in part by the National Natural Science Foundation of China (62271172), in part by the China Postdoctoral Science Foundation (2023M733615).

	

	
	
	{~~~~}
	
	
	\noindent\textbf{References}
	\bibliography{reference}

	
	
	
	
	
	
	
\end{document}